%% file: acl2023.tex
\pdfoutput=1
\documentclass[11pt]{article}

\usepackage[]{ACL2023}

\usepackage{times}
\usepackage{latexsym}
\usepackage{amsmath}

\usepackage{algorithm}
\usepackage{algpseudocode}

\newcommand\sFor[2]{ \For{#1}#2\EndFor} % snappy version of \ForAll...\EndFor
\usepackage{graphicx} 
\usepackage{geometry}
\usepackage{subfig}
\usepackage{tabularx}
\usepackage{booktabs}
\usepackage{multirow}
\usepackage{amsfonts}

\usepackage[T1]{fontenc}
\usepackage[utf8]{inputenc}

\usepackage{microtype}

\usepackage{inconsolata}

\usepackage{xspace}

\DeclareMathOperator*{\argmax}{arg\,max}

\newcommand{\vQ}{\mathbf{Q}}
\newcommand{\vK}{\mathbf{K}}
\newcommand{\vS}{\mathbf{S}}
\newcommand{\vP}{\mathbf{P}}
\newcommand{\vO}{\mathbf{O}}
\newcommand{\vR}{\mathbf{R}}
\algnewcommand{\LineComment}[1]{\State \(\triangleright\) #1}

\newcommand{\mname}{\textsc{NaCl}\xspace}
\newcommand{\mnameA}{\textsc{Proxy-Tokens Eviction}\xspace}
\newcommand{\mnameB}{\textsc{Random Eviction}\xspace}

\newcommand\blfootnote[1]{%
\begingroup
\renewcommand\thefootnote{}\footnote{#1}%
\addtocounter{footnote}{-1}%
\endgroup
}

\title{\mname: A General and Effective KV Cache Eviction Framework for LLMs at Inference Time}

\author{
    Yilong Chen$^{1,2*}$,
    ~Guoxia Wang$^{3*}$,
    ~Junyuan Shang$^{3^\ddagger}$,
    ~\textbf{Shiyao Cui}$^1$,
    ~Zhenyu Zhang$^3$,\\
    ~\textbf{Tingwen Liu}$^{1,2^\dagger}$\textbf{,}
    ~\textbf{Shuohuan Wang}$^3$\textbf{,}~\textbf{Yu Sun}$^3$\textbf{,}~\textbf{Dianhai Yu}$^3$\textbf{,}~\textbf{Hua Wu}$^3$ \\ 
    \normalsize $^1$ Institute of Information Engineering, Chinese Academy of Sciences\\
    \normalsize $^2$ School of Cyber Security, University of Chinese Academy of Sciences\\
    \normalsize $^3$ Baidu Inc.\\
    \small \{\texttt{chenyilong, cuishiyao, liutingwen\}@iie.ac.cn} \\
    \small \{\texttt{wangguoxia, shangjunyuan, zhangzhenyu07, wangshuohuan, sunyu02\}@baidu.com}
}

\begin{document}
\maketitle
\begin{abstract}
% \blfootnote{Work done during Yilong Chen’s internship at Baidu.}
Large Language Models (LLMs) have ignited an innovative surge of AI applications, marking a new era of exciting possibilities equipped with extended context windows. However, hosting these models is cost-prohibitive mainly due to the extensive memory consumption of KV Cache involving long-context modeling. Despite several works proposing to evict unnecessary tokens from the KV Cache, most of them rely on the biased local statistics of accumulated attention scores and report performance using unconvincing metric like perplexity on inadequate short-text evaluation. In this paper, we propose \mname, a general framework for long-context KV cache eviction that achieves more optimal and efficient eviction in a single operation during the encoding phase. Due to \mname's efficiency, we combine more accurate attention score statistics in \mnameA with the diversified random eviction strategy of \mnameB, aiming to alleviate the issue of attention bias and enhance the robustness in maintaining pivotal tokens for long-context modeling tasks. Notably, our method significantly improves the performance on short- and long-text tasks by 80\% and 76\% respectively, reducing KV Cache by up to $5\times$ with over 95\% performance maintenance. The code is available at \url{https://github.com/PaddlePaddle/Research/tree/master/NLP/ACL2024-NACL}.

\end{abstract}

% \blfootnote{$^*$Equal contribution $^\dagger$Corresponding author $^\ddagger$ Project lead}
\blfootnote{$^*$Equal contribution. Work done at Baidu Inc. }
\blfootnote{$^\dagger$Corresponding author. $^\ddagger$ Project lead.}
% \blfootnote{}

\section{Introduction}
Large Language Models (LLMs) with longer context window~\cite{touvron2023llama, xiong2023effective, jiang2023mistral, claude, GPT-4} have emerged recently for better conducting long conversations,
summarizing long documents, or debugging code at the repository level~\cite{bai2023longbench}. However, their deployment is costly and infeasible on fixed memory hardware, mainly due to the surprisingly large memory consumption of \textit{KV Cache} mechanism. For instance, a 7 billion-parameter model
with an input batch size of 4 and a sequence length of 32k results in 64GB of KV cache, $4.7\times$ larger than the model weights.

\begin{figure}[t]
    \centering
    \includegraphics[width=0.9\linewidth]{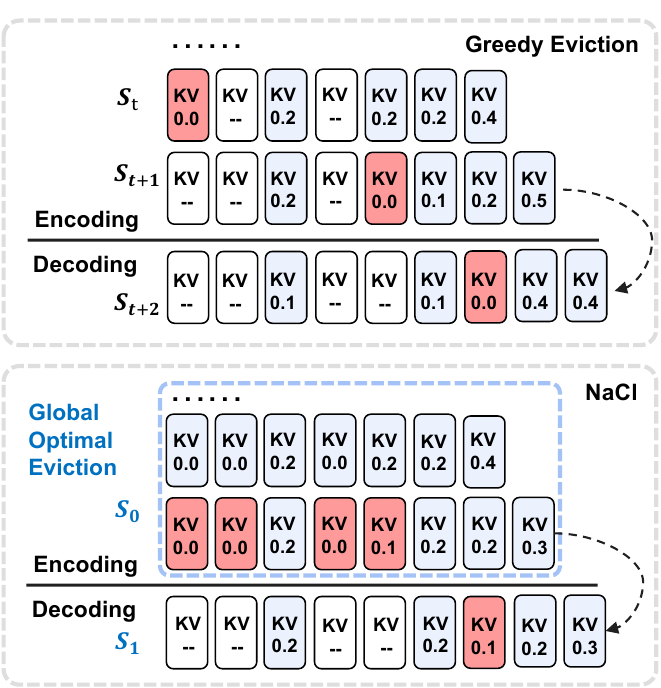}
    \caption{Traditional eviction algorithms perform step-by-step greedy search for tokens for eviction. Our framework searches globally for tokens within a chunk and then performs one single eviction.}
    \label{fig:intro-framework}
    \vspace{-0.25cm}
\end{figure}

To mitigate the
pressure on the scarce GPU memory from using KV cache, a number of studies~\cite{h2o, liu2023scissorhands, ge2023model, sink} have explored sparsity among Transformer attention blocks to evict unnecessary tokens from the KV cache. For instance, H2O~\cite{h2o} utilized the local statistics of accumulated attention scores to retain a balance of recent and heavy hitter tokens during generation. Window Attention based methods~\cite{sink} proposed to keep the initial tokens which is proven vital for generation fluency. This line of work reduced the memory footprint of KV cache for efficient inference with negligible loss in generation quality. In addition, the above methods do not require costly retraining which is more suitable for current open-sourced LLMs~\cite{touvron2023llama, jiang2023mistral}, compared to those that need specific attention mechanism adaptation~\cite{beltagy2020longformer, kitaev2020reformer, shazeer2019fast, ainslie2023gqa}.

However, we argue that the performance reported in the above methods is over-optimistic, as the evaluation metric and tested benchmark is not sufficient. LLMs may fail in real-life long-context modeling tasks~\cite{bai2023longbench}, though they can achieve low language modeling perplexity which is untrustworthy used as the golden metric in current studies~\cite{sink, lminfinite}. Furthermore, the local statistics of accumulated attention score is observed to be biased (see Fig.~\ref{fig:observation}), especially in long context input, meaning that it should be carefully used as the only strategy for measuring the importance of tokens.

To fill the gap, we propose \mname, a ge\underline{n}er\underline{a}l and effective KV \underline{c}ache eviction framework to unleash the power of L\underline{L}Ms for long-context modeling with limited memory budgets. \mname specifically formulates the eviction task in encoding phase which is different from the commonly used one-token-in one-token out eviction procedure in generation phase. In encoding phase, the eviction can be effectively implemented to apply only once on the whole input by progressively evicting KV caches layer by layer. The one-eviction formulation benefits current eviction policies in a more efficient and optimal way, as multiple costly eviction operations can be combined, then the global statistics of attention scores can be utilized. 

Based on the above formulation, we present \mnameA which exploits the global statistics of attention scores gathered from proxy tokens for eviction. In practise, the proxy tokens can be selected from the question input, commonly located at the end of a long text. Intuitively, these proxy tokens are more capable of retaining the task-specific tokens in KV cache. As a result, \mnameA alleviates the \textit{attention bias problem} (see Sec.~\ref{sec:obsevation}) occurred in methods using local statistics~\cite{h2o, msrnn} or task-irrelevant proxy tokens~\cite{liu2023scissorhands}. 

However, \mnameA also relies heavily on the statistic of attention scores which may be untrustworthy in long-context input. Thus, we incorporate \mnameB, a random eviction policy, into \mnameA. \mnameB randomly samples tokens to evict from the probability distribution in \mnameA with different seed on attention heads and layers. This diversified randomness enhances the model's robustness to maintain potentially important tokens in long text generation.

We conducted extensive experiments on a single NVIDIA A100 (80GB) GPU on representative open-sourced LLMs: LLaMA2-base, LLaMA2-Chat~\cite{touvron2023llama}, and evaluated them on both short- and long-text modeling tasks from lm-eval-harness~\cite{eval-harness} and LongBench~\cite{bai2023longbench}. The experiments show that \mname performs KV cache eviction efficiently with negligible degradation on model quality (i.e., saving the inference memory usage of KV cache by up to $5\times$ with over 95\% maintenance). Specifically, \mname achieve 80\% and 75\% performance improvement on short- and long- text modeling tasks, respectively, with 50\% KV cache reduction, compared to current eviction methods.

\section{Related Work}
\paragraph{Efficient Inference with Limited KV Cache Budgets} emerged for reducing the prominent inference bottleneck caused by KV cache, particularly for long content input. A series of methods~\cite{h2o, ge2023model, liu2023scissorhands, msrnn} explored the sparsity among Transformer's attention block, then evicted unnecessary tokens from KV Cache for efficient inference. For instance, H2O~\cite{h2o} retained a balance of recent and heavy hitter tokens with the highest accumulated attention scores throughout the sequence. Scissorhands~\cite{liu2023scissorhands} sequentially predicted the potentially pivotal tokens with the attention score above average within a history window. Some method~\cite{ge2023model} further applied costly eviction policy selection for better performance. However, the above methods relied heavily on the attention score with local statistics which may be sub-optimal in long-context tasks~\cite{longchat2023, bai2023longbench}.

Meanwhile, some efforts have been made to utilize a learnable mechanism to determine necessary tokens during inference~\cite{anagnostidis2023dynamic}, or converting the traditional multi-head attention(MHA)~\cite{vaswani2017attention} to multi-query attention (MQA)~\cite{shazeer2019fast} or group-query attention (GQA)~\cite{ainslie2023gqa}. However, these methods involve additional training, while \mname focuses on the inference phase without resource-intensive training. 

\paragraph{Efficient Transformers}~\cite{tay2020efficient} have been extensively explored~\cite{sparse, kitaev2020reformer, bigbird, beltagy2020longformer, transformer_xl, erniedoc, rmt, chevalier2023adapting} to address the self-attention operation which scales quadratically with the sequence length. For instance, Sparse Transformer~\cite{sparse} uses a dilated sliding window the reduces the attention complexity. Longformer~\cite{beltagy2020longformer} and Bigbird~\cite{bigbird} reduced the complexity of self-attention by combining random, window and global attention. Recurrence Transformers~\cite{transformer_xl} maintain a memory bank of past KV cache to process the long text in segments. However, the above methods either trade off model quality or require re-training of models, but often failed in achieving memory saving and wall-clock speedup at inference time~\cite{dao2022flashattention}.

\paragraph{Length Extrapolation} enabled language models to generalize beyond the context window they were trained on. A recent line of research~\cite{chen2023extending, peng2023yarn, liu2023scaling} focuses on adapting relative positional embedding~\cite{su2024roformer} widely used in current Foundation models~\cite{touvron2023llama, jiang2023mistral} for context window extension. Attention Sink~\cite{sink} and LM-Infinite~\cite{lminfinite} further exploited the initial tokens to recover the performance of window attention for infinite-length inputs. However, the ability of these methods tested using metric like perplexity is over-optimistic for long context tasks~\cite{longchat2023, bai2023longbench}.

\section{Problem Formulation}

This section defines a two-phased approach for efficient KV cache management during LLM inference, tailored for scenarios with limited KV cache budgets.\vspace{+0.15cm}

\paragraph{Eviction Policy} We defined the eviction policy $F_{\text{score}}: S_{t}^{i} \leftarrow S_{t-1}^i, \text{subject to } |S_{t}^i|=|S_{t-1}^i| \leq \mathcal{C}$ where the scoring function $F_{\text{score}}$ assigns low scores to unnecessary tokens for eviction, such that the pre-define KV cache budget $\mathcal{C}$ is maintained. $S_{t}^i$ denote the indices set of retained tokens in KV cache at $t$-th time step and $i$-th transformer layer.

\paragraph{Encoding Phase Eviction} The model processes the input prompts, $x_{\text{prompt}}^i=[x_1^i, \ldots, x_p^i] \in \mathbb{R}^{p \times d}$ , to compute the initial key cache $\mathcal{K}_0^i=x_{\text{prompt}}^i  W_K^i \in \mathbb{R}^{p \times d}$ and value cache $\mathcal{V}_0^i=x_{\text{prompt}}^i  W_V^i \in \mathbb{R}^{p \times d}$, where $p$ denotes the encoding prompt length, $W_K^i, W_V^i \in  \mathbb{R}^{d \times d} $ represent the key and value projection weight at layer $i$ with hidden dimension $d$, respectively. The attention scores $\mathbf{A}_{\text{prompt}}^i \in \mathbb{R}^{p \times p}$ are computed as $\frac{(x_{\text{prompt}}^i W_Q^i) \cdot {(x_{\text{prompt}}^i W_K^i)}^T}{\sqrt{d}}$ where $W_Q^i \in  \mathbb{R}^{d \times d} $ represent the query projection weight. The eviction in encoding phase is defined as follows:
\begin{equation*}
    S_{\text{encoding}}^i = F_{\text{score}}(\mathbf{A}_{\text{prompt}}^i, \mathcal{C})
\end{equation*}
then, the initial KV cache can be updated $\mathcal{K}_0^i,\mathcal{V}_0^i \leftarrow \mathcal{K}_{S_{\text{encoding}}^i}, \mathcal{V}_{S_{\text{encoding}}^i}$ for the later usage in generation phase.

\paragraph{Generation Phase Eviction} Denote the generated tokens' input to $i$-th layer as $x_{\text{decoding}}^i=[z_1^i,\ldots,z_T^i] \in \mathbb{R}^{T \times d}$. The Generation phase updates the KV cache with each new token generation. Given the time step $t$ and layer $i$, key and value cache is updated as $\mathcal{K}_{t}^i = [\mathcal{K}_{t-1}^i, z_t^i \cdot W_K^i]$, $\mathcal{V}_{t}^i = [\mathcal{V}_{t-1}^i, z_t^i \cdot W_V^i]$, respectively. The attention scores $\mathbf{A}_{t}^i \in \mathbb{R}^{1 \times |\mathcal{K}_{t}^i|}$ are computed as $\frac{(z_{t}^i W_Q^i) \cdot {\mathcal{K}_{t}^i}^T}{\sqrt{d}}$. The eviction in generation phase is defined as follows:
\begin{equation*}
    S_t^i = F_{\text{score}}(\mathbf{A}_{t}^i, S_{t-1}, \mathcal{C})
\end{equation*}
where the KV cache are updated $\frac{T}{m}$ times following $\mathcal{K}_t^i,\mathcal{V}_t^i \leftarrow \mathcal{K}_{S_t^i}, \mathcal{V}_{S_t^i}$ at every $m$ time steps.

% The $\text{Update}$ function ensures the cache size remains within $\mathcal{C}$ by selectively adding new entries or pruning existing ones.
% $K_{S_t, *}\left(\in \mathbb{R}^{k \times d}\right)$ denotes a sub-matrix of $K$ which selects $S_t$ rows from $K$.
To note that, recent works formulated the encoding phase eviction the same as the one in generation phase which require \textit{step-by-step} evictions, resulting in computational overhead. In contrast, we formulate the eviction to perform only once during the encoding phase and $\frac{T}{m}$ times during the generation phase. Generally, the condition \( T \ll p \) is readily satisfied in long text scenarios, allowing $\frac{T}{m}$ to be approximated as a constant order of magnitude. Consequently, the overall time complexity is reduced from \( \mathcal{O}(p + T) \) to \( \mathcal{O}(1) \). This also allows the eviction policy in a global optimal manner comparing to those greedy algorithm that couples the input window size with the KV cache budget.

\begin{figure*}[!ht]
\centering
\subfloat[H2O]{
\includegraphics[width=0.228\textwidth]{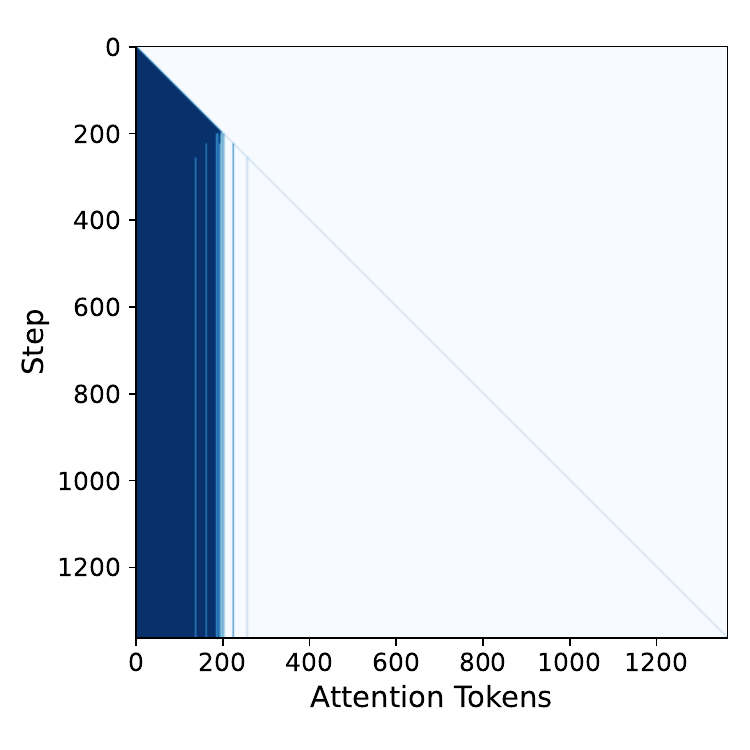}\label{fig:h2o}
}
\subfloat[MSRNN]{
\includegraphics[width=0.22\textwidth]{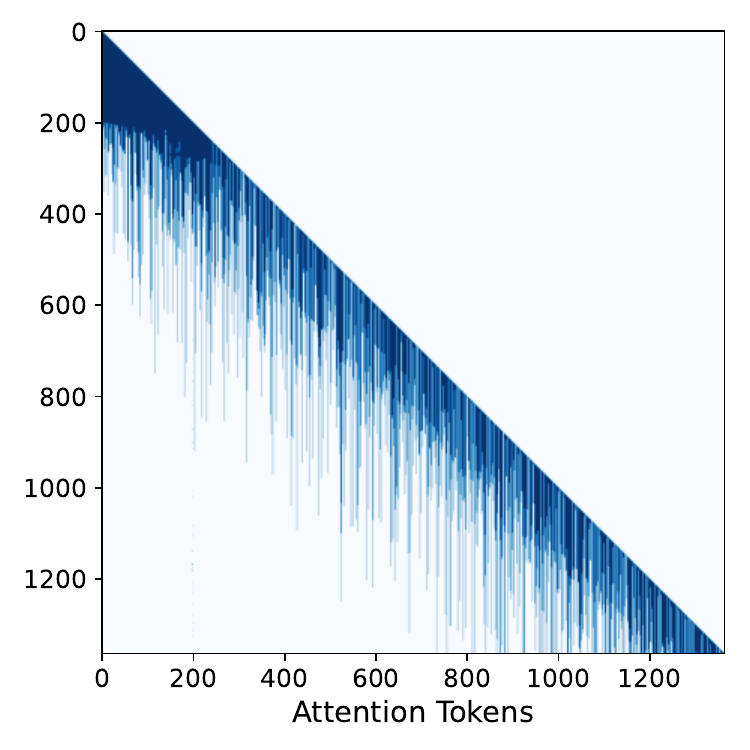}\label{fig:msrnn}
}
\subfloat[\mname]{
\includegraphics[width=0.22\textwidth]{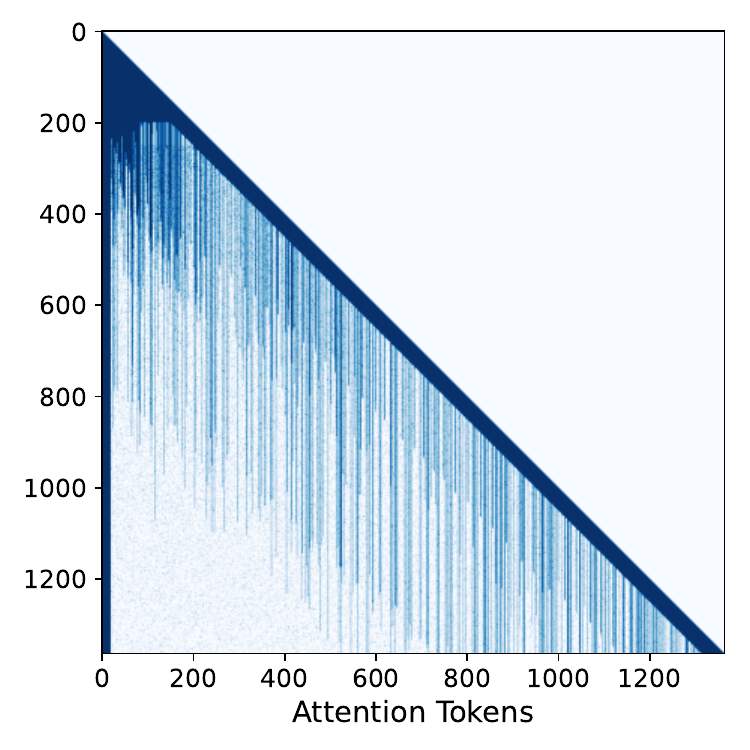}\label{fig:nacl}}
\subfloat[Attention Scores in Token Indices]{
\includegraphics[width=0.28\textwidth]{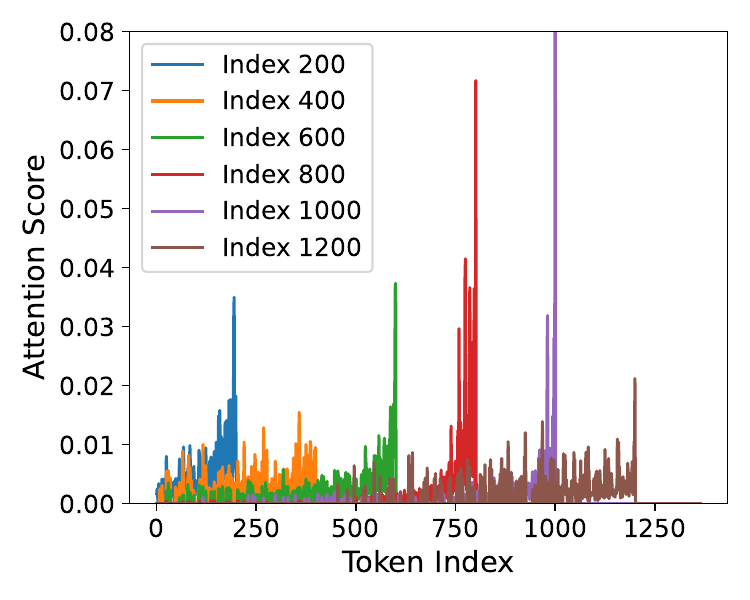}\label{fig:attnscore}}

\caption{Attention score bias in eviction policy. The darker color in Fig. (a,b,c) shows the retained tokens.}
\label{fig:observation}
\vspace{-0.25cm}
\end{figure*}
\section{Observation}\label{sec:obsevation}

We present two experimental findings by rethinking previous eviction methods that inspire the design of \mname.

\paragraph{Rethinking Evaluation Metrics for  Long-text Eviction Strategy} Current metrics such as the perplexity (PPL) fall short in capturing the nuances of model performance in long-text scenarios, revealing a gap between evaluation practices and real-world applications (see, Tab.~\ref{tab:short-text-exp}). Evaluations predominantly utilize datasets with short texts, inadequately representing the complexities and challenges of processing and understanding long-text input. The emphasis on textual fluency leads to a notable bias: the method~\cite{sink}, though claiming for infinite input, fails in tasks (see, Tab.~\ref{tab:short-text-exp},~\ref{tab:long-text-exp}) which requires the ability to generate accurately. This inspires us to re-evaluate current methods on both short- and long- text modeling tasks demands on comprehension and generation capabilities.

\paragraph{Rethinking Attention Scores to Retain Pivotal Tokens}\textit{Attention bias problem} refers to the phenomenon where, at each step of generation, attention scores are higher within the tokens directly preceding the current token, while comparatively diminished for all others. In Fig.~\ref{fig:observation} (a) and (b), the \textit{attention bias problem} is observed, leading to an overemphasis on either initial tokens~\cite{h2o} or recent tokens~\cite{msrnn}, overlooking those potentially pivotal tokens in longer context. Furthermore, the attention score distribution become flattened with the increase in text length (see Fig.~\ref{fig:observation} (d)), which may be less capable of accurately identifying important tokens.  Normalization can solve this problem to some extent, but as stated in the H2O~\cite{h2o} , the effect is not optimal. This inspires us to reform the attention-based eviction methods to be less bias and more robust in long-context modeling tasks.

\section{\mname}
In this section, we present a hybrid KV cache eviction policy in \mname, including the \mnameA in Sec.~\ref{sec:Proxy-Tokens} and \mnameB in Sec.~\ref{sec:multi-dice}.

\begin{figure*}
    \centering
    \includegraphics[width=0.9\linewidth]{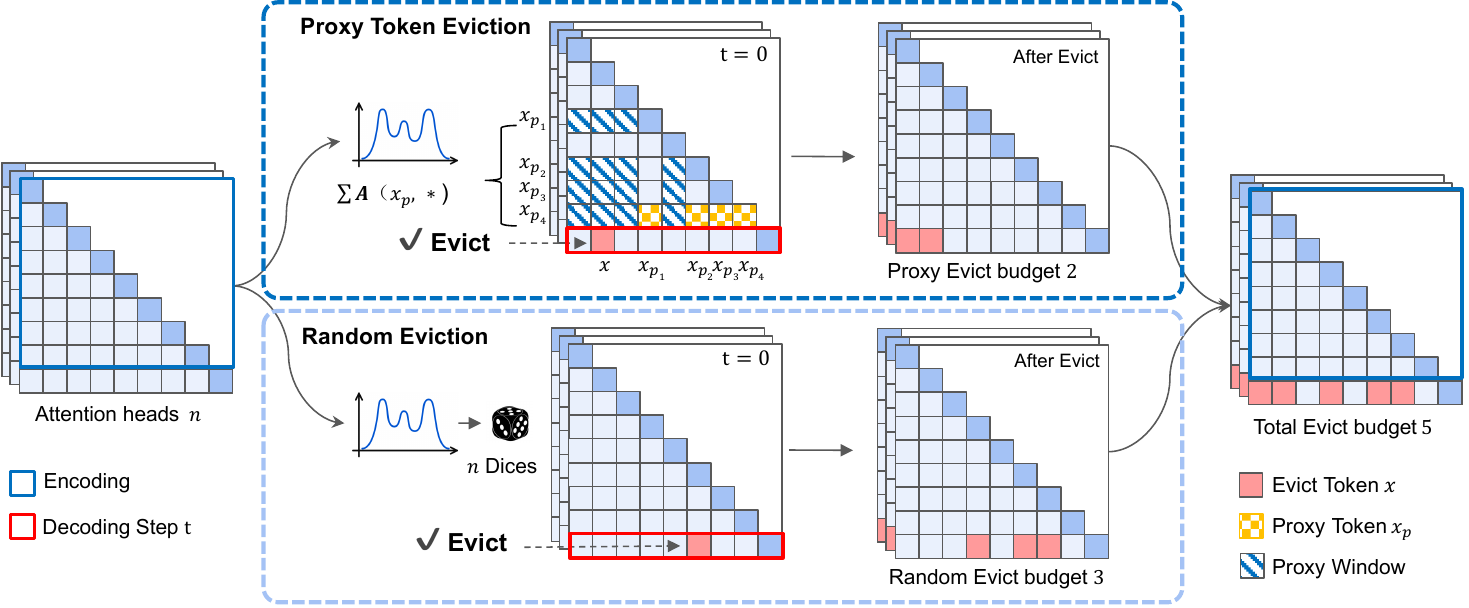}
    \caption{\mname consists of a hybrid eviction policy by incorporating \mnameB into \mnameA. \mnameA utilizes proxy tokens for more accurate eviction, while \mnameB performs head-wise sampling from the scoring function of \mnameA to enhance the robustness.}
    \label{fig:enter-label}
    \vspace{-0.25cm}
\end{figure*}

\subsection{Eviction based on Proxy Tokens}\label{sec:Proxy-Tokens}

Based on previous observations, the current $F_{\text {score }}$ of accumulating attention scores effectively identifies important tokens but suffers from significant bias. We attribute this to the excessive redundant information in the process of scoring tokens. 

We discovered that when calculating the attention for a given token \(x\) (i.e. tokens may need to be evicted), only a mere fraction of  tokens \(x_p\) (i.e. proxy tokens) are responsible for yielding the most precise outcomes during the computation of the token score. Hence, we introduce the proxy tokens hypothesis: within the input \(x_{\text{prompt}}\), there exists a subset called proxy tokens \(\mathcal{P}  \in x_{\text{prompt}}\), which precisely estimate the importance of tokens. Scoring function $F_{\text {score }}$ can be instantiated as:
\begin{equation}
\small
 F_{\text {score }}\left(\mathbf{A}, \mathcal{C} \right)= \sum_{x_p \in \mathcal{P} } \text{Softmax}\left(\mathbf{A}(x_{p}, *) \right) 
    % F_{\text {score }}\left(\mathbf{A}, \mathcal{C} \right)=
    % \sum_{x_p \in \mathcal{P} }
    % \text{Softmax}\left(\mathbf{A}(x,x_p) \right)
\nonumber
\end{equation}

The $F_{\text {score }}$, calculated by reducing the attention score matrix column-wise using the proxy tokens subset \(\mathcal{P} \), provides the most precise measurement of a token's importance during the eviction process. We can validate the significance of proxy tokens using a straightforward approach. When the proxy token is set to the universal set, our method is equivalent to H2O~\cite{h2o}, introducing redundant information that degrades the quality of eviction. When the proxy token is set to only the current token, our method can be equivalent to MSRNN~\cite{msrnn}, neglecting substantial information, thus reducing the accuracy of eviction.

Due to the progressively flattened distribution of attention scores with the increase in text length, the pre-defined threshold for sampling \(\mathcal{C}_p\) results in lack of generalizability for long text tasks. Therefore, we model the KV cache eviction as an optimization problem, aiming to find a set \(S_t\) that maximizes the function \(F_{\text{score}}\), while satisfying the constraint \(|S_t| = \mathcal{C}_p\). 
\begin{equation}
    \small
 S_t \leftarrow (\argmax_{S_t \subset R }  \sum_{x \in S_t} F_{\text{score}}(\mathbf{A}, \mathcal{C}_p)) \cup P
 \nonumber
\end{equation}
where $R = x_{\text{prompt}} \backslash \mathcal{P}$ as the proxy tokens are retained by default. In practise, the proxy tokens tend to be chosen at the end of the input where the user's question with more task-specific information is located in. The choice of Proxy Token can be based on task orientation, which allows our approach to be flexibly adapted to various application scenarios. For more information, please refer to the Appx.~\ref{app:how proxy tokens loc}.

\subsection{Eviction based on Random Possibility Sampling}\label{sec:multi-dice}

% 为什么引入随机性
% 什么情况下引入随机性好，什么情况下引入随机性不好
Eviction algorithms commonly rely on the attention scores which may be biased or lack robustness in capturing critical information throughout the generation process. Herein, we introduce a simple and effective eviction policy which incorporates the randomness into the decision-making process of the attention mechanism. By randomly sampling from a probability distribution, our method aims to enhance the model's ability to recover and maintain important information that might otherwise be lost. 

In detail, we construct the probability distribution from \(F_{\text{random}}\) . \(F_{\text{random}}\) can signify each candidate token's relative significance in long text generation, and the probability \(P_{\text {prompt }}\) is determined as follows:
% (vW_Q) \cdot {(vW_K)}^T {\sqrt{d_k}}^{-1}
\begin{equation}
\small
    P_{\text {prompt }} = \text{Softmax}\left(F_{\text {random }}(\mathbf{A}_{\text {prompt }} , \mathcal{C}_r)\right)
\nonumber
\end{equation}
where $P_{\text {prompt}}$ allows the non-deterministic selection of pivotal tokens. Through this probabilistic lens, our model casts the dice, diversifying its focus and increasing the chances of preserving essential information across the span of long texts. Thus, we present \mnameB with budget $\mathcal{C}_r$:
\begin{equation}
    S_{\text{random}} \sim P_{\text{prompt}}, \quad |S_{\text{random}}|=\mathcal{C}_r
    \nonumber
\end{equation}
In practise, $P_{\text{random}}$ can be based on the normalized distribution Softmax($F_{\text{score}}$), then the complexity is $\mathcal{O}(|x_\text{prompt}|)$ dominated by the softmax operation. 

Finally, \mname effectively combines \mnameA and \mnameB, applying an efficient one-eviction strategy under the KV Cache budget $\mathcal{C}=\mathcal{C}_p + \mathcal{C}_r$, shown in the following Algorithm~\ref{algo:algorithmic}. Our method is compatible with FlashAttention-2 (see Appx.~\ref{app:fa impl}) to minimize memory and computational overhead, helping models to be efficiently deployed in long text tasks.

\begin{algorithm}[t!]
\small
\caption{\mname Algorithm}
\begin{algorithmic}[1]
\State  \text{Total Cache budget} $\mathcal{C}$ ($\mathcal{C} = \mathcal{C}_{\text{p}} + \mathcal{C}_{\text{r}}$), Proxy-Token Eviction Cache budget $\mathcal{C}_{\text{p}}$, Random Eviction Cache budget $\mathcal{C}_{\text{r}} $, \text{Proxy tokens} $\mathcal{P}$, KV Cache $\mathcal{K}, \mathcal{V}$ 
\Function{Encoding}{Prompts}
\For{Every Layer-$i$ in LLMs}
    \For{Every Attention Head-$n$}
            \State $W_Q^{i,n}, W_K^{i,n}, W_V^{i,n} \in \mathbb{R}^{d \times d}$
            \State $ \mathbf{A} \leftarrow (x_{\text{prompt}} W_Q^{i,n}) \cdot {(x_{\text{prompt}} W_K^{i,n})}^T {\sqrt{d}}^{-1}$
            \State $F_{\text {score}}=
                \sum_{x_p \in \mathcal{P} }
                \text{Softmax}\left(\mathbf{A}(x_p, *) \right)$
            \State $R \leftarrow  x_{\text{prompt}} \backslash P^{i,n}$
            \State $u_{\text{score}} \leftarrow (\max \sum_{x \in R} F_{\text{score}}(\mathbf{A}, \mathcal{C}_{\text{p}})) 
            \cup \mathcal{P}^{i,n}$
            \State $u_{\text{random}} \sim \text{Softmax}\left(F_{\text {random}}(\mathbf{A}_{\text{prompt}}) , \mathcal{C}_{\text{r}})\right)$
            \State $S_{\text{encoding}}^{i, n} \leftarrow u_{\text{score}}  \cup u_{\text{random}}$
    \EndFor
\EndFor
\EndFunction

\Function{Generation}{$S_{\text{encoding}}$, Max Length}
\State $m \leftarrow$ eviction interval
\State $z_{0} \leftarrow$ last prompt token
\State $S_0 \leftarrow S_{\text{encoding}}$

 \For{t $\in\{1, ..., \text{Max Length}\}$}
    \For{Every Layer-$i$ in LLMs}
        \For{Every Attention Head-$n$}
            \State $\mathcal{K}_{t-1}^{i, n} \leftarrow \mathcal{K}_{S_{t-1}^{i, n}}, \mathcal{V}_{t-1}^{i, n} \leftarrow \mathcal{V}_{S_{t-1}^{i, n}}$
            \State $\mathcal{K}_{t}^{i, n} \leftarrow [\mathcal{K}_{t-1}^{i, n}, z_t^i \cdot W_K^{i, n}]$
            \State $\mathcal{V}_{t}^{i, n} \leftarrow [\mathcal{V}_{t-1}^{i, n}, z_t^i \cdot W_V^{i, n}]$
            \State $ \mathbf{A} =(z_{t} W_Q^{i,n}) \cdot {\mathcal{K}_{t}^{i, n}}^T {\sqrt{d}}^{-1}$
            \If{$t \bmod m = 0$} % If 语句，需要和EndIf对应
            \State $S_t^{i, n} \leftarrow \text{Eviction}(\mathbf{A}, \mathcal{C})$
            % \Comment{Ref: Line7-10.}
            \LineComment{Ref: Line7-10.}
            % \Comment{Line7-10.}
            \EndIf
        \EndFor
    \EndFor
    \State $\boldsymbol{z}_t \leftarrow$ sample from LLM prediction
\EndFor
\EndFunction
\end{algorithmic}
\label{algo:algorithmic}
\end{algorithm}

\section{Experiments}

\subsection{Setup}
\begin{table*}[ht]
\centering
\small
\begin{tabularx}{\textwidth}{lccccccccccc}
\toprule
 \textbf{Model}& \textbf{PiQA}& \textbf{COPA}& \textbf{Open.}& \textbf{Wino.}& \textbf{SciQ}  &\textbf{ARC-E}&\textbf{ARC-C}&\textbf{Average}  & \textbf{$\Delta$} & \textbf{log PPL} \\
\midrule
\# of tokens (5-Shot) &319& 118& 97& 160& 508 & 296 &239& $-$ & $-$ & $-$ \\ 
\midrule
Full cache & 78.8 & 83.0 & 44.8 & 73.7 & 80.9 & 78.8 & 50.8 & 64.6 & $-$ & 3.8 \\
\midrule
Attention Sink (20\%) & 54.0 & 55.0 & 30.2 & 49.1 & 22.3 & 25.4 & 23.0 & 35.9 &$-$28.7 & 6.4 \\
H2O (20\%) & 77.6 & \textbf{81.0} & 41.0 & 67.0 & 75.8  & 70.4 & 44.0 & 60.3 &$-$4.3 & 4.0 \\
MSRNN(20\%) & 77.6 & 78.0 & 43.0 & 67.8 & 76.5 & 71.6 & 45.3 & 60.6& $-$4.0& 4.0\\
% Scissorhands(20\%) & 77.6 & 78.0 & 43.0 & 67.8 & 76.5 & 71.6 & 45.3 & 60.6& \\
\mname (20\%) & \textbf{77.9} & 79.0 & \textbf{43.8} & \textbf{71.5} & \textbf{80.0} & \textbf{74.9} & \textbf{48.8} & \textbf{63.8}& \textbf{$-$0.8} & 4.0\\
\midrule
\# of tokens (25-Shot) & 1014 & 501 & 559 & 689 & 2540 & 1480 & 1195 & $-$ & $-$ & $-$\\ 
\midrule
Full cache & 59.3 & 87.0 & 47.2 & 75.5 & 11.1 & 67.6 & 31.2 & 53.8& $-$ & 3.2 \\ 
\midrule
Attention Sink (20\%)& 50.4 & 47.0 & 29.0 & 46.6 & \textbf{11.1} & 25.6 & 22.5 & 33.2& $-$20.6& 7.9\\
H2O (20\%)& \textbf{59.2} & 86.0 & 44.6 & 73.8 & 10.5 & 66.0 & 30.0 & 52.8& $-$1.0& 3.3 \\  
MSRNN (20\%)& 58.9 & 86.0 & 44.8 & \textbf{73.9} & 10.7 & 65.9 & 30.6 & 52.9& $-$0.9& 3.3 \\
% Scissorhands (20\%)& 58.9 & 86.0 & 44.8 & 73.9 & 10.7 & 65.9 & 30.6 &&\\
\mname (20\%)& 58.9& \textbf{87.0} & \textbf{45.6} & 73.6 &\textbf{11.1} & \textbf{66.1} & \textbf{31.4} & \textbf{53.2} & \textbf{$-$0.6} & 3.2 \\
 \bottomrule
\end{tabularx}
\caption{N-shot evaluation of eviction strategies on short text tasks on LLaMA2-7B-base.}
\label{tab:short-text-exp}
% \vspace{-0.5cm}
\end{table*}

% and Yarn-Mistral-7B~\cite{jiang2023mistral}
\paragraph{Objective} We aim to provide experimental evidence for three key research questions: \textbf{1}. Whether there are advantages in performance and task generalization of \mname over other eviction methods. \textbf{2}. How the two eviction policys in \mname affect the final functionality, and by what combination can we achieve optimal results. \textbf{3}. What is the rationale behind \mname for superior results?

\paragraph{Models and Tasks} We use the family of decoder-only Transformers: LLaMA2-7B-base, LLaMA2-7B-Chat~\cite{touvron2023llama} to evaluate the effectiveness of \mname. To evaluate the few-shot learning ability, we sample seven tasks from the popular benchmark (lm-eval-harness~\cite{eval-harness}):  PiQA~\cite{PiQA}, COPA~\cite{COPA}, OpenBookQA~\cite{OpenbookQA}, Winogrande~\cite{Winogrande}, SciQA~\cite{SciQA}, ARC-E and ARC-C~\cite{ARC}. In the long text scenario, we took seven tasks from Longbench~\cite{bai2023longbench}: PassageRetrieval-Zh, PassageRetrieval-En, RepoBench-P~\cite{RepoBench}, HotpotQA~\cite{HotpotQA}, NarrativeQA~\cite{NarrativeqQA}, TriviaQA~\cite{TriviaQA}, QMSum~\cite{QMSum}. We report perplexity computed on the OpenBookQA dataset as a measure of the model's generation ability in generalized domains. We conduct our experiments on a single NVIDIA A100 80GB GPU. Results were averaged over various seeds to ensure reliability.

\paragraph{Baselines} We consider four representative eviction methods:

\textbf{\textbullet \text{ } Attention Sink}~\cite{sink} keeps the initial and recent tokens for infinite-length text processing.

\textbf{\textbullet \text{ } H2O}~\cite{h2o} firstly proposes utilizing the summation of attention scores for greedy eviction, which achieves fair results which serves as our main baseline.

\textbf{\textbullet \text{ } MSRNN}~\cite{msrnn} considers the current token's attention score for eviction.

\textbf{\textbullet \text{ } Scissorhands}~\cite{liu2023scissorhands} increments the counter within a history window for low score token eviction.

\begin{figure}[t]
    \centering
    \includegraphics[width=0.7\linewidth]{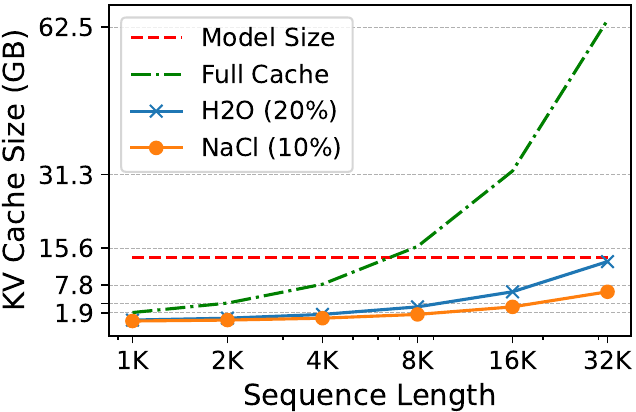}
    \caption{The memory usage of KV Cache with respect to the sequence length in the setting of comparable downstreaming performance between \mname and other methods.}
    \label{fig:cache}
\vspace{-0.25cm}
\end{figure}

\begin{table*}[ht]
\setlength{\tabcolsep}{5.2pt}
\centering
\small
\begin{tabularx}{\textwidth}{lcccccccccc}
\toprule
 \textbf{Method}& \textbf{PR-Zh}& \textbf{PR-En}& \textbf{Repo-P}& \textbf{HotpotQA}& \textbf{NarQA}  &\textbf{TriviaQA}&\textbf{QMSum}&\textbf{Average}  &\textbf{$\Delta$}\\
\midrule
Full Cache & 8.0 & 10.1 & 52.3 & 27.7 & 18.6 & 83.3 & 20.6 & 31.5 & -- \\
\midrule
Scissorhands (30\%) & 4.0 & 2.6 & 51.8 & 17.2 & 12.5 & 79.2  & 19.6 & 26.7 & $-$4.8 \\
H2O (30\%) & 3.7 & 5.0 & 50.9 & 27.1 & 15.5 & 81.6 & 20.2 & 29.1 & $-$2.4 \\
MSRNN (30\%) & 5.5 & 4.5 & 50.3 & 26.1 & \textbf{16.4} & 82.9 & 20.9 & 29.5 & $-$2.0\\
\mname(30\%) & \textbf{6.8} & \textbf{9.0} & \textbf{52.5} & \textbf{27.9} & \textbf{16.4} & \textbf{83.1} & \textbf{21.5} & \textbf{31.0 } & \textbf{$-$0.5}\\
\midrule
Scissorhands (20\%) & 0.5 & 43.0 & 44.9 & 11.4 & 6.7 & 68.9 & 16.3 & 21.9 & $-$9.6 \\
H2O (20\%) & 4.2 & 4.5 & 49.4 & 24.5 & 15.2 & \textbf{82.8} & 19.8 & 28.6 & $-$2.9 \\
MSRNN(20\%) & 4.5 & 4.5 & 49.0 & 23.9 & 14.7 & 82.5 & 20.4 & 28.5 & $-$3.0\\
\mname (20\%) & \textbf{7.0} & \textbf{9.4} & \textbf{51.6} & \textbf{27.2} & \textbf{17.1} & 82.5 & \textbf{20.8} & \textbf{30.8} & \textbf{$-$0.7}\\
\midrule
Scissorhands (10\%) & 0.0 & 3.7 & 27.6 & 4.4 & 2.8 & 52.8 & 12.9 & 14.9 & $-$16.6\\
H2O (10\%)& 4.9 & 3.5 & 48.1 & 22.8 & 13.6 & 79.4 & 19.6 & 27.4 & $-$4.1 \\
MSRNN (10\%)& 4.0 & 3.0 & 47.9 & 23.4 & 13.4 & 80.8 & 19.6 & 27.5 & $-$4.0 \\
\mname(10\%)& \textbf{6.8} & \textbf{7.0} & \textbf{49.2} &\textbf{25.5} & \textbf{15.0} & \textbf{81.7} & \textbf{20.4} & \textbf{29.4} & \textbf{$-$2.1}\\
 \bottomrule
\end{tabularx}
\caption{Evaluation of eviction strategies on long text tasks with 4k-length on LLaMA2-7B-Chat.}
\label{tab:long-text-exp}
% \vspace{-0.5cm}
\end{table*}

\subsection{Result}

\input{tables/ablation}

% 不同budget大小下的性能 短文本
\paragraph{Short-Text Performance} Our experimental analysis shows the effectiveness of \mname in managing KV cache under constrained memory budgets while maintaining high performance across various short-text benchmarks. Firstly, \mname demonstrated superior performance in comparison to the baseline eviction methods with minimal performance degradation. In the 5-shot setting, \mname achieved an average score of 63.8\% points, nearly matching the full cache performance of 64.6\% points and significantly outperforming H2O by 3.5\% points. 
% For example, in the 5-shot setting, \mname outperformed H2O and MSRNN in datasets such as Winogrande and SciQ, demonstrating its adeptness at selecting relevant information across diverse domains.
Moreover, \mname exhibited consistent improvements across most datasets relative to previous methods, affirming its robust generalization and practical applicability. In the 25-shot setting, although there was a performance dip across all methods due to the increased complexity and information redundancy, \mname still showed remarkable resilience. Notably, it matched or slightly outperformed the full cache setup in certain datasets, such as maintaining a 87.0\% point score on COPA, identical to the full cache performance. This illustrates that \mname not only manages to select pertinent information effectively but also mitigates the impact of redundant data, enhancing the model's robustness.
% 不同budget大小下的性能 长文本

\paragraph{Long-text Performance} \mname achieves 80\% memory usage reduction with only mere 0.7\% point decrease with respect to the average accuracy in Tab.~\ref{tab:long-text-exp}. Fig.~\ref{fig:ablation-studies} (Left) shows that \mname is possible to achieve $3\times$ more reduction in KV Cache while maintaining comparable performance to baselines. Additionally, we observed the stable performance of \mname under different budget, while others' fluctuate.
In HotpotQA and QMSum, \mname (30\%) even surpassed the performance without KV cache eviction by 0.2\% and 0.9\% points, respectively. For challenging passkey retrieval tasks, H2O and MSRNN with the attention bias towards initial and recent tokens fails in retaining the pivotal passkey located in the middle of the long input. In contrast, \mname demonstrates stable and superior performance in different budgets setting, that only missed 2 passkeys in PR-Zh and PR-En comparing to the model in full cache setting. This remarkable achievement highlights \mname's ability to retain essential information, avoiding the pitfalls of redundant data and thereby bolstering the model's robustness in processing complex long texts.

\subsection{Ablation Studies}

\begin{figure*}[!t]
\centering
\includegraphics[width=0.32\textwidth]{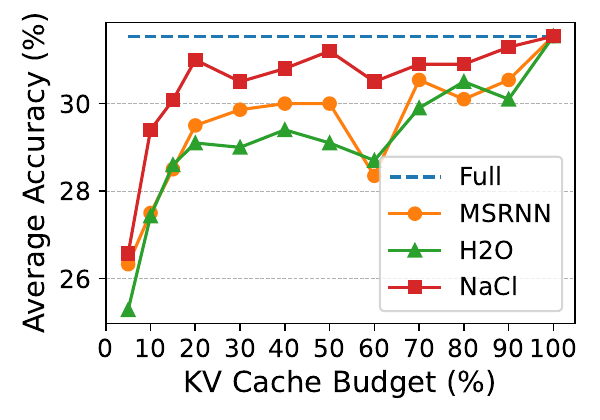}
\includegraphics[width=0.32\textwidth]{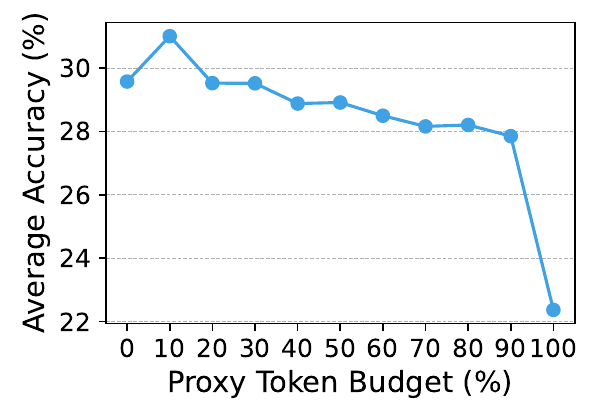}
\includegraphics[width=0.32\textwidth]{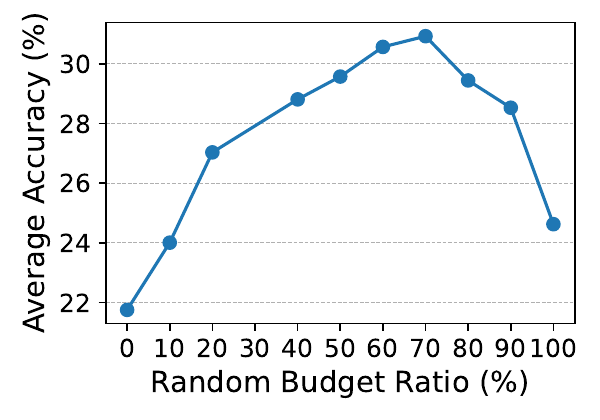}
\caption{The average accuracy is reported with different KV Cache budget (Left), Proxy tokens budget (Middle), and Random budget ratio (Right).}
\label{fig:ablation-studies}
\vspace{-0.25cm}
\end{figure*}

%  Set the budegt of \mnameA and \mnameB to $a$ and $r$ respectively. We have $k=f+r$.

% headwise和非headwise

% 初期增长：当随机预算占比从10%增加到40%时，平均准确率先是略有下降（从21.664下降到21.252），然后上升到21.588。这表明在随机预算占比较低时，准确率对预算比例的变化较为敏感，且有可能通过适当增加随机预算比例来提高准确率。
% 中期下降：随着随机预算占比的进一步增加，从40%增加到80%，平均准确率呈现明显的下降趋势。这个阶段的下降较为均匀，表明在这个区间内，增加随机预算占比对于提升准确率是不利的。
% 后期急剧下降：特别是当随机预算占比从80%增加到90%时，平均准确率从20.65下降到15.434，出现了明显的跳跃性下降。这暗示在随机预算占比非常高时（接近全部预算），可能会对准确率产生极大的负面影响。

% win的影响

\paragraph{The Effect of \mnameA}\label{sec.proxyeff} Proxy tokens play an important role in finding pivotal tokens. The performance degradation (see Tab.~\ref{tab:ablation}) is significant when removing this policy. In Fig.~\ref{fig:ablation-studies} (Middle), we report the impact of proxy token budget on the average accuracy as a proportion of the text length. In extreme cases, such as 0\% and 100\% proxy token budget, the method degenerates into two special cases: MSRNN~\cite{msrnn} and H2O~\cite{h2o}, respectively. The suboptimal performance with 0\% proxy token budget suggests that the unsufficiency of a single current token for determining the pivotal tokens. However, excessive abuse of proxy token budget up to 100\% will introduce redundant information leading to decline in performance. In practise, we suggest the budgets for proxy tokens $\sim$10\% for better performance.
% With proxy token budget between 0.05-0.2 and 0.8-0.9, we can obtain a more accurate and smoother scoring function.

\paragraph{The Effect of \mnameB} As shown in Tab.~\ref{tab:ablation}, \mnameB obtained a performance gain of 9\% points. In Fig.~\ref{fig:ablation-studies} (Right), increasing the random budget from 10\% to 70\% results in an average performance improvement of 2.25\% to 8.17\% points over 0\% random budget. The performance peaks at 70\% budgets, demonstrating the necessity of the \mnameB. However, when the random budget's proportion increases from 90\% to 100\%, a noticeable performance decline occurs, highlighting the importance to combine the Attention-score-based, \mnameA.
% 基于不同的random概率采样

\paragraph{The Choice of Sampling Distributions for \mnameB} In Tab.~\ref{tab:ablation}, the experimental results demonstrate that sampling based on global statistical attention scores outperforms those based on uniform distributions in terms of performance. This indicates that attention scores can also provide a more informative reference for randomness in specific contexts.
% In contrast, other random methods primarily rely on assumed distribution shapes rather than the actual distribution characteristics of the data.
% 一次性驱逐，滑窗驱逐

\paragraph{The Effect of Global Eviction} The one-eviction formulation in \mname enhanced the average performance of 1.4\% points in Tab.~\ref{tab:ablation}. Compared to the greedy algorithm, our approach reduces complexity while giving more consideration to global information. Furthermore, our algorithm exhibits greater simplicity and directness in its engineering implementation.
% headwise和非headwise

\paragraph{The Effect of Head-wise Token Eviction} The results in Tab.~\ref{tab:ablation} show a gradual decline in the algorithm's effectiveness as the strategies become more uniform. The algorithm performs best when each head adopts a completely different strategy. The diversity of strategies leads to improved generalization, preserving information across a broader spectrum of dimensions.

% \begin{figure}
%     \centering
%     \includegraphics[width=1\linewidth]{fig/win.png}
%     \caption{Enter Caption}
%     \label{fig:enter-label}
% \end{figure}
% \textbf{Proxy token budget:}
% In Figure X, we report the impact of the proportion of nearby protection \(c\) relative to text length on the average accuracy of \mnameA. Introducing \(c\) in the range of 0.025 to 0.05 effectively enhances performance, confirming our hypothesis of uniform sampling and the importance of local tokens' multiplicity in the generation process proposed in the Methods section. Further increasing \(c\) beyond this range results in a marginal decrease in performance. Therefore, we set \(c\) at 0.05 in our optimal implementation.

\subsection{Analysis}
\paragraph{Memory Usage of KV Cache} In Fig.~\ref{fig:cache}, we report the memory reduction of KV cache for LLaMA2-7B with respect to the sequence length with a fixed batch size of 4 in bf16 precision. At the same accuracy as H2O (20\%), our method alleviates the linearly growing KV cache to 10\% of the original size, significantly reducing the memory footprint of the KV cache for all model sizes. The memory usage reduction is more significant on long texts, effectively alleviating the memory bottleneck problem of long text reasoning.

% 采样注意力的可视化
\paragraph{Interpretable Analysis of Eviction Results} Fig.~\ref{fig:observation} shows the attention score matrix where the darker area represents the retained tokens after eviction. Compared to previous methods, the proxy token guides \mname to sample the middle tokens more evenly, and protects the initial and recent tokens in the meantime. The head-wise randomness enables maintaining more context information, thereby enhancing the robustness of \mname.

\paragraph{Why Head-wise Eviction Matters} From a probabilistic perspective, it is basically impossible for a token to be evicted in the head-wise eviction setting. Taking LLaMA-7B with 32-layers (number of layers $l$) and 32-heads (number of heads $h$) as an example, the probability of a token retained in least one head's KV cache is   which equals 99.92\% when the KV Cache budget $\mathcal{C}=20\%$. Even in a severe KV Cache budget setting like $\mathcal{C}=1\%$, the probability that the information of a token is retained in at least one layer is $1 - (\mathcal{C}^{\text{$h$}})^{\text{$l$}}$ which is larger than 99.99\%.

\section{Conclusion}
In this paper, we focus on the accuracy, robustness, and reliability of evaluation for KV cache eviction algorithms deployed in LLMs for processing long texts. We introduce \mname, a novel approach that combines \mnameA and \mnameB for KV cache eviction strategies, significantly reducing memory usage during model inference without the need for training. We model the eviction problem as a combinatorial optimization issue, where \mnameA provides eviction references based on importance, and \mnameB enhances information richness and robustness through headwise and layerwise composite sampling. Through extensive evaluation, we demonstrate that \mname can significantly improve cache eviction strategies, reduce inference memory costs, and minimize the impact on the LLM’s ability to handle complex tasks.

\section*{Limitations}
Our approach presents two main limitations: First, due to constraints on resources, our method has not been extensively tested across various large-scale language models, especially for different lengths and even ultra-long texts. However, based on our current comprehensive experimental conclusions, we believe \mname can be extended to more application scenarios. In addition, we introduced the utilization of proxy tokens in \mnameA for identifying pivotal tokens, yet the selection of proxy tokens primarily relies on observations and experience. Determining proxy tokens from the model adaptively and accurately presents a challenge, which we deem worthy of further research.
% proxy
% 没有搜索最优的超参数

\section*{Ethics Statement}
In this research, we employ open-source data and technologies, significantly reducing privacy concerns. Our innovative approach is geared towards understanding model contexts and boosting inference efficiency, with the aim of developing accessible and highly efficient models for extended contexts. This strategy is anticipated to propel the openness of NLP technology and its practical implementation in diverse applications. Importantly, our method is designed to be independent of the training process, ensuring it does not perpetuate or introduce biases into models. By focusing on cutting-edge and resource-efficient methodologies, our work contributes to making AI more open and automated, pushing the envelope in artificial intelligence while ensuring the benefits of our advancements are widely accessible and applicable across various domains, marking a step towards a more inclusive and automated AI future.

\section*{Acknowledgments}
We thank the anonymous reviewers for their insightful comments and constructive suggestions. We would like to thank Yinqi Yang, Jiawei Sheng, Xinhua Zhang, Shicheng Wang, Chuanyu Tang and members of the IIE KDsec group for their valuable feedback and discussions. We thank Siming Wu for the implementation of Reduce Attention Scores CUDA kernel. Work done during Yilong Chen's internship at Baidu Inc. This research is supported by the National Key Research and Development Program of China (Grant No.2021ZD0110501 and Grant No.2021YFB3100600) and the Youth Innovation Promotion Association of CAS (Grant No.2021153).

% \section*{Acknowledgements}

\bibliography{anthology,custom}
\bibliographystyle{acl_natbib}

\appendix

% \newpage
% \newpage

\section{Appendix}\label{sec:appendix}

\subsection{Sparsity in Attention Cache}
Inspired by previous literature on the existence of attentional sparsity in self-attentive heads, we delve into the sparsity of attention during the generation of LLMs. Given a normalised attention score matrix calculated by the softmax function applied to the dot product of the query matrix $Q$ and the key matrix $K$, represented as $A =\text{Softmax}(QK^\top)$, the attention mechanism allocates weights to different elements in the input sequence, reflecting their relative importance. Thus if the attention score of a token is low, it means that it has little influence on the process, and therefore we base our sparsification on the threshold to quantify sparsity. The sparsity percentage for a given threshold t is calculated as:
\[{\text{Sparsity}}(t,i) = \frac{ \sum_{j=1}^{N} \mathbf{1}\left(|A_{ij}| < t\right)}{N}\]
where \(N\) is the dimension of the attention matrix, \(\text{Softmax}(QK^\top)_{ij}\) represents the attention weight between the \(i^{th}\) and \(j^{th}\) elements, and \(\mathbf{1}(\cdot)\) is an indicator function that returns 1 if the condition is true and 0 otherwise. This formula calculates the proportion of attention weights that are considered negligible or insignificant for different sparsity thresholds, thus providing a multi-faceted view of attention distribution's sparsity across the model.

\begin{figure}[t]
    \centering
    \includegraphics[width=\linewidth]{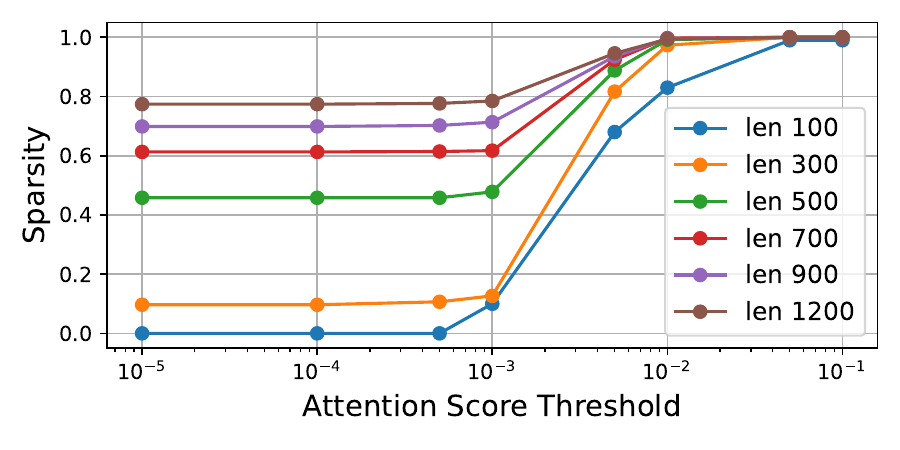 }
    \caption{Attention weights sparsity under different thresholds and sequence length.}
    \label{fig:sparsity}
\end{figure}

Fig.~\ref{fig:sparsity} shows the change of Sparsity under different thresholds, it can be seen that under different threshold values, the sparsity of A gradually increases with the sequence length, when the sequence length is 1200, the sparsity is more stable at 0.78. This means that 22\% of the tokens are the dominant factor in the computation process.

\subsection{Attention Score Function in Eviction}\label{app:attn score}
\begin{figure}[!t]
    \centering
    \includegraphics[width=\linewidth]{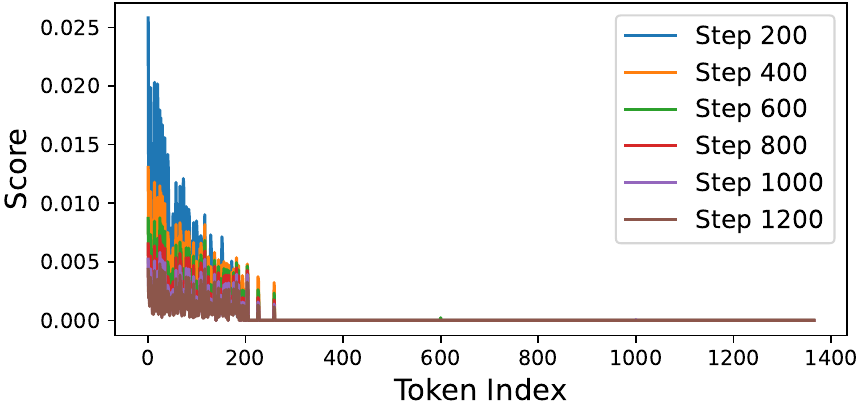}
    \caption{H2O Eviction Function Score with Step}
    \label{fig:h2o_fs}
    \vspace{-0.5cm}
\end{figure}

\begin{figure}[!t]
    \centering
    \includegraphics[width=\linewidth]{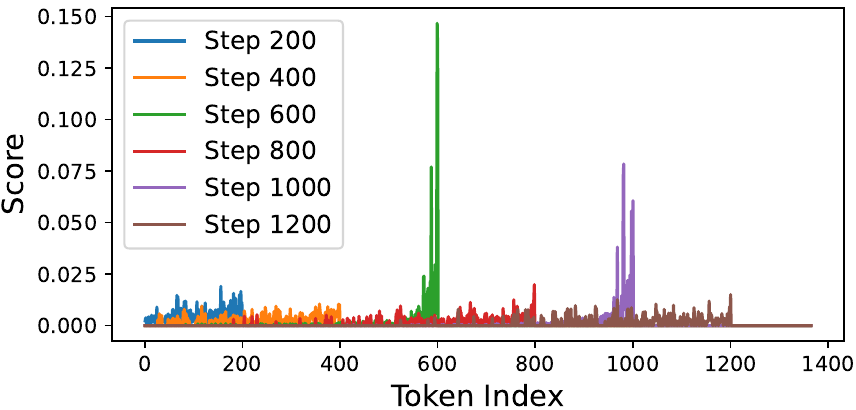}
    \caption{MSRNN Eviction Function Score with Step}
    \label{fig:msrnn_fs}
    \vspace{-0.5cm}
\end{figure}

\begin{figure}[!t]
    \centering
    \includegraphics[width=\linewidth]{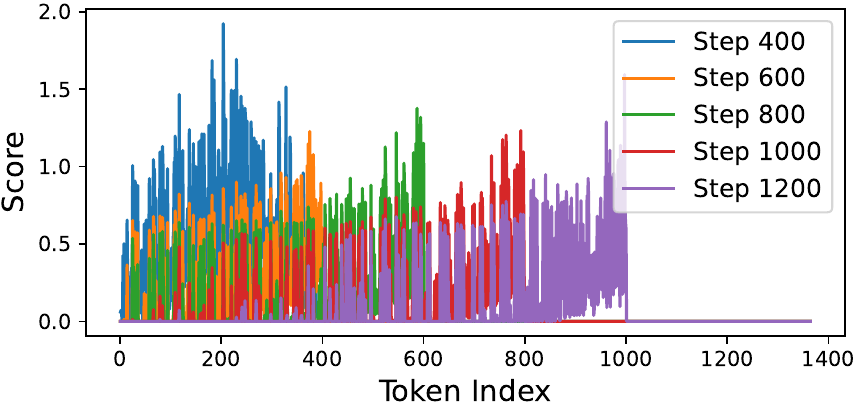}
    \caption{NACL Eviction Function Score with Step}
    \label{fig:nacl_fs}
    \vspace{-0.5cm}
\end{figure}

In this section we show the importance scores of token in the contexts for H2O (see Fig.~\ref{fig:h2o_fs}), MSRNN (see Fig.~\ref{fig:msrnn_fs}) and NACL (see Fig.~\ref{fig:nacl_fs}) during different steps. The importance score function of H2O assigns larger scores to tokens in the front part, and MSRNN assigns larger scores to tokens in close proximity.NACL evenly distributes importance scores over longer contexts, so that both distant and close tokens will have a chance to be sampled to be retained.


\subsection{The Influence of Proxy Token Locations on NACL’s Proxy Eviction Strategy}\label{app:how proxy tokens loc}
In Proxy Token Eviction, we need to find the most accurate tokens in the sequence that can calculate the importance score of the token.The core of this problem is how to establish a judgment criterion of whether a token is important or not, and this judgment criterion determines how to select the proxy token set.
Intuitively, we believe that whether a token in a sequence is important or not is determined by the task the model is about to accomplish. We refer to this task of the model's input as the user's question. In practical applications, we can usually separate the user's question, so that we can place it at the end of input to maximize the performance of the Proxy Eviction Strategy. When we are unaware of the position of the user's question, we primarily utilize proxy tokens to protect the beginning and end of the sequence, as in most cases, these positions contain crucial information about the generation. Even if the proxy token fails to include any information related to the question, our method can be considered as an improved version of MSRNN~\cite{msrnn}. By introducing the proxy token, we regularize the distribution of the Important scores. We also enhance the robustness of preserving intermediate information by combining it with the Random Sample strategy. Based on the aforementioned combination strategies, NACL demonstrates significant performance improvements in terms of results.

\subsection{Why \mname ues Random Eviction?}\label{app:why random}

\begin{figure}[ht]
    \centering
    \includegraphics[width=1\linewidth]{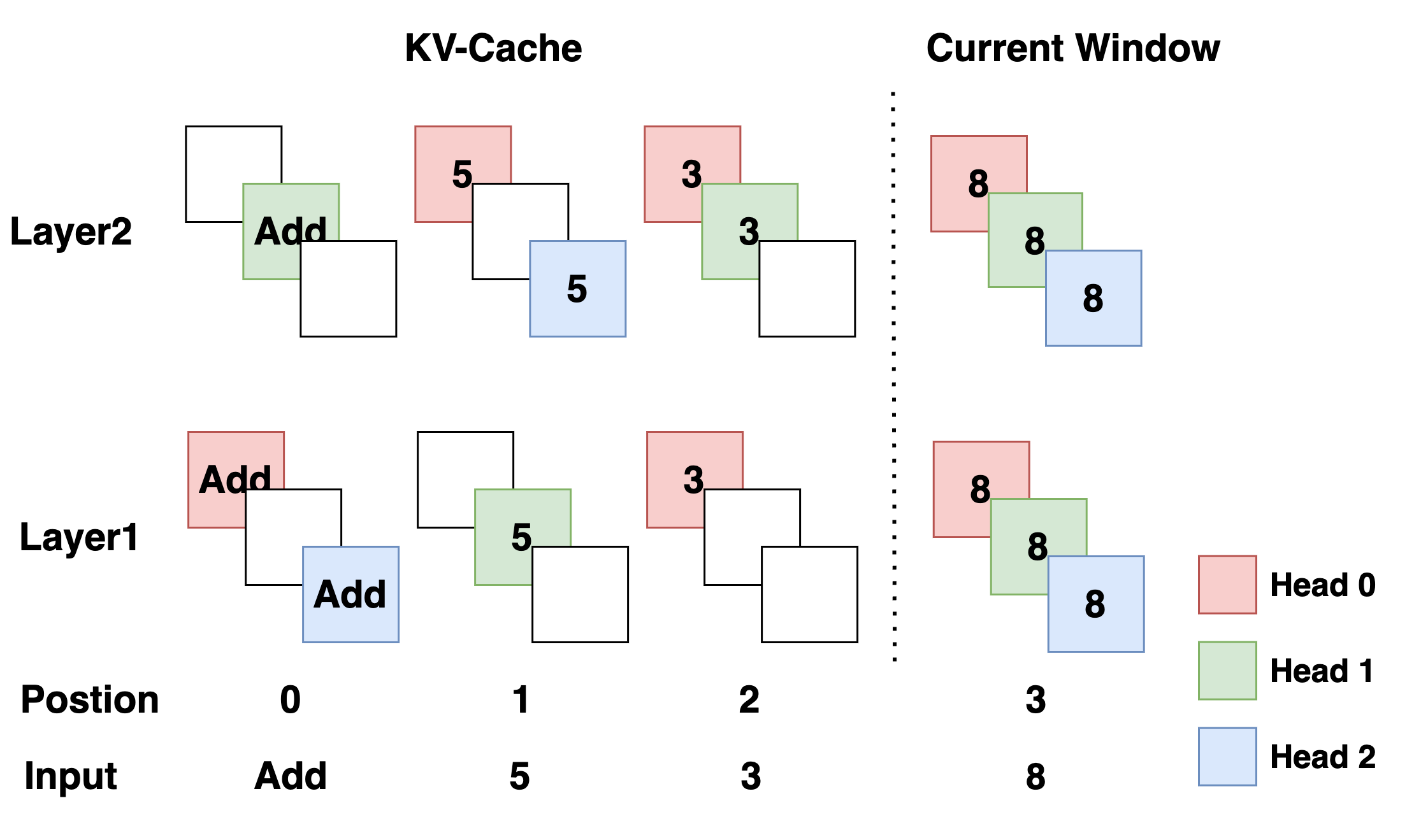}
    \caption{Through random sampling, these layers and heads can access a broader range of tokens, thereby casting a wider net to capture information.}
    \label{fig:example-random}
\end{figure}
We demonstrate a case of combinatorial optimization based on randomness in Figure~\ref{fig:example-random}. Within the KV cache, the loss of any information can lead to a misunderstanding of the current information. In previous methods, a broader spectrum of information that might otherwise be overlooked due to the uniformity and potential bias of score-based eviction methods. This randomness of diversification ensures that even information that is not prominently featured by self-attention mechanisms is given a chance to be integrated at deeper levels of the model~\cite{transformer_xl}. Consequently, this approach facilitates the model's ability to capture and process a more varied set of token interactions, enhancing its overall performance by reducing the risk of losing vital information.

\subsection{Algorithmic Complexity Analysis}
In long-text scenarios, the length of the generated text during the generation phase is often much shorter than the length of the input text. Assuming the length of the input text during the encoding phase is \(p\) and the length of the output text during the decoding phase is \(T\), where \(T \ll p\). Given the complexity of the \(F_{\text{score}}\) function as \(O(p)\), and applying a \textit{step-by-step} eviction algorithm on data of length \(p\) results in a complexity of \(O(p^2)\), making it impractical and costly for real-world applications.

Therefore, in long-text scenarios, we employ a one-time optimal eviction algorithm, where we calculate the optimal eviction strategy \(S_{enocding}\) in a one-time computation during the encoding phase. Since the number of tokens generated is negligible compared to the total, we apply the eviction strategy \(S_i\) at this stage as well. In comparison to the greedy algorithm that evicts based on \(S_{i-1}\) while retaining the same budget, our method can globally optimize to find the best eviction strategy. Moreover, we can decrease the time complexity from \(O(p^2)\) to \(O(p)\), making the algorithm straightforward and effective in engineering applications.

\subsection{Hyperparameters}
Here, we provide the hyperparameters for allocating the ratio of KV cache bugets for the hybrid eviction policy used in our experiments in Tab. \ref{tab:hyperp}.
\begin{table*}[t]
    \centering
    \begin{tabular}{cccccc}
    \toprule
         Budget&  Protect Proxy &  No-protect Proxy & Proxy-Tokens Eviction& Random Eviction\\
    \midrule
         10\%&  1\%&  3\%&  2\%&7\%\\
         20\%&  2\%&  18\%&  6\%&12\%\\
 30\%& 1.5\%& 20\%& 10.5\%&18\%\\
    \bottomrule
    \end{tabular}
    \caption{The allocation of the KV cache budget ratio for Protect Proxy, \mnameA and \mnameB in \mname.}
    \label{tab:hyperp}
\end{table*}

\subsection{Reduce Attention Scores with FlashAttention-2} \label{app:fa impl}

We have implemented NACL on 128k long-text inference and made it compatible with Flash Attention. There are two implementations below, both of which can be directly used with Flash Attention.

\paragraph{Re-Computation of the Attention Score:} We utilize $\vQ_{\mathcal{P}}$ and $\vK$ to calculate the required attention scores during the encoding phase, separate from Flash Attention. Since the proxy tokens set is very small, only a small portion of the attention score needs to be re-computed, thus the additional overhead is insignificant. According to experimental results, on a 128k context, evicting 20\% while maintaining a stable 15GB of memory usage does not affect the inference speed.

\paragraph{Implementation of Reduce Attention Scores Kernel:} The forward computation of FlashAttention-2~\cite{dao2022flashattention} returns the log-sum-exp (Logsumexp) for each row. Leveraging this Logsumexp, we can recompute the attention scores matrix in the manner described in the backward computation of FlashAttention-2. Subsequently, we perform a column-wise summation to obtain the reduced attention scores, as outlined in Algorithm~\ref{alg:reduce_attn_scores}.

\begin{figure*}[ht]
  \centering
  \begin{minipage}{0.9\textwidth}
    \begin{algorithm}[H]
      % \footnotesize
      % \scriptsize
      \caption{\label{alg:reduce_attn_scores} Reduce Attention Scores with FlashAttention-2}
      \begin{algorithmic}[1]
        \Require Matrices $\vQ \in \mathbb{R}^{N_q \times d}, \vK \in \mathbb{R}^{N_k \times d}$ in HBM, vector Logsumexp $L \in \mathbb{R}^{N_q}$ in HBM, block sizes $B_c$, $B_r$.
        \State Divide $\vQ$ into $T_r = \left\lceil\frac{N_q}{B_r} \right\rceil$ blocks $\vQ_1, \dots, \vQ_{T_r}$ of size $B_r \times d$ each,
        and divide $\vK$ in to $T_c = \left\lceil \frac{N_k}{B_c} \right\rceil$ blocks $\vK_1, \dots, \vK_{T_c}$, of size $B_c \times d$ each.
        \State Divide $L$ into $T_r$ blocks $L_i, \dots, L_{T_r}$ of size $B_r$ each.
        \State Initialize the output $\vO = (0)_{N_k}$ in HBM and divide it into $T_c$ blocks $\vO_1, \dots, \vO_{T_c}$ of size $B_c$ each.
        \sFor{$1 \le j \le T_c$}{
          \State Load $\vK_j$ from HBM to on-chip SRAM.
          \State Initialize $\vR_j = (0)_{B_c}$ on Register.
          \sFor{$1 \le i \le T_r$}{
            \State Load $\vQ_i, L_i$ from HBM to on-chip SRAM.
            \State On chip, compute $\vS_{i}^{(j)} = \vQ_i \vK_j^T \in \mathbb{R}^{B_r \times B_c}$.
            \State On chip, compute $\vP_{i}^{(j)} = \exp(\vS_{ij} - L_{i}) \in \mathbb{R}^{B_r \times B_c}$.
            \State On chip, compute
            $\vR_j \leftarrow \vR_j + Reduce(\vP_{i}^{(j)}) \in \mathbb{R}^{B_c}$.
            }
            % \ENDFor
        \State amtomicAdd($\vO_j, \vR_j$).}
        % \ENDFor
        \State Return $\vO$.
      \end{algorithmic}
    \end{algorithm}
  \end{minipage}
\end{figure*}

\end{document}

%% file: tables/ablation.tex
\begin{table}[ht]
    \centering
    
    \resizebox{0.48\textwidth}{!}{%
    \begin{tabular}{@{}lccc@{}}
    \toprule
     & \textbf{Short-Text ACC.}& \textbf{Long-Text ACC.}\\ \midrule
    \mname & 63.8 & 30.8 \\ \midrule
    $-$ Eviction w. Proxy Tokens& 35.7 ($-$28.1)&  24.8 ($-$6.0) \\ 
 $-$ Random Eviction& 62.6 ($-$1.2) & 21.7 ($-$9.2) \\
 $-$ Probability Sampling&63.0 ($-$0.8)  &29.7 ($-$1.1)\\
 $-$ Global Eviction& 62.5 ($-$1.3) &29.3 ($-$1.5) \\
 $-$ Head-wise Eviction& 61.7 ($-$2.1) &28.1 ($-$2.7)\\
 \bottomrule
    \end{tabular}
    }
    \caption{Ablation study at 20\% budget's eviction. We report the average accuracy of short- and long-text tasks.} 
    \label{tab:ablation}
\vspace{-0.5cm}
\end{table}

% \begin{table*}[13]{ht}{0.48\textwidth}
%     \vspace{-1.2em}
%     \centering
%     \small    
%     \caption{
%         A summary of pre-training datasets used by \ours{} and other models.
%     } 
%     \resizebox{0.48\textwidth}{!}{%
%     \setlength{\tabcolsep}{3pt}
    
%     \begin{tabular}{lccccc}
%     \toprule
%     \textbf{Method}& \textbf{PR} &Repo-p &HotpotQA &NarQA& \textbf{Average}\\ \midrule
%     \manme&  & & && \\ %
%     - \manmeA&  & & && \\ 
%     \midrule
%     - \manmeB&  & & && \\
%     - Nearby protection&  & & && \\
%     - random sampling&  & & && \\
%     - Global eviction&  & & && \\
%     - Headwise eviction&  & & && \\
%     \ours &  & & && \\
    
%     \bottomrule 
%     \end{tabular}
% bg     }

%  $-$ Nearby protection& & \\

%     \label{tab:baseline_config}
% \end{table*}